\documentclass[runningheads]{llncs}
\pdfoutput=1
\usepackage{graphicx}

\usepackage{amsmath,amssymb} 
\usepackage{color}

\usepackage{hyperref}

\graphicspath{{./images/}}

\usepackage[linesnumbered,ruled,vlined]{algorithm2e} 

\SetCommentSty{mycommfont}

\usepackage{authblk}
\usepackage[hang,flushmargin]{footmisc}

 \newcommand{\chapternote}[1]{{%
  \let\thempfn\relax
  \footnotetext[0]{\text{#1}}
}}

\begin{document}
\title{Task-Oriented Hand Motion Retargeting for Dexterous Manipulation Imitation}

\titlerunning{Task-Oriented Hand Motion Retargeting}

\author{Dafni Antotsiou \and
Guillermo Garcia-Hernando \and
Tae-Kyun Kim}

\authorrunning{Dafni Antotsiou \and
Guillermo Garcia-Hernando \and
Tae-Kyun Kim}

\institute{Imperial College London\\ \{d.antotsiou17, g.garcia-hernando, tk.kim\}@imperial.ac.uk} 

\maketitle             
\vspace{-5mm}
\begin{abstract}
Human hand actions are quite complex, especially when they involve object manipulation, mainly due to the high dimensionality of the hand and the vast action space that entails.
Imitating those actions with dexterous hand models involves different important and challenging steps: acquiring human hand information, retargeting it to a hand model, and learning a policy from acquired data.
In this work, we capture the hand information by using a state-of-the-art hand pose estimator. We tackle the retargeting problem from the hand pose to a 29 DoF hand model by combining inverse kinematics and PSO with a task objective optimisation. This objective encourages the virtual hand to accomplish the manipulation task, relieving the effect of the estimator's noise and the domain gap. Our approach leads to a better success rate in the grasping task compared to our inverse kinematics baseline, allowing us to record successful human demonstrations. 
Furthermore, we used these demonstrations to learn a policy network using generative adversarial imitation learning (GAIL) that is able to autonomously grasp an object in the virtual space.
\vspace{-2mm}

\keywords{hand pose estimation, motion retargeting, PSO, anthropomorphic hand model, imitation learning, GAIL.}
\end{abstract}

\section{Introduction}
\chapternote{Project webpage: \url{https://daphneantotsiou.github.io/task-oriented-retargeting.html}}
Learning to perform human-like tasks is an important goal of artificial intelligence. Achieving this goal though presents many challenges, predominantly adjusting the tasks to the agent's (i.e. robots) architecture and inferring intention about the task's desired outcome. This work is interested in the imitation of tasks performed by the human hand - such as grasping - using a dexterous anthropomorphic hand model.

There are two main difficulties when tackling this problem. First is the interpretation of the human motion to the agent's environment, which is called retargeting. Second is the inference of the tasks' objective and the ability to perform them in the agent's environment. One way of achieving this goal is through imitation learning, which involves using human demonstrations that the agent attempts to imitate \cite{hussein2017imitation}. That can become difficult, though, when the agent is a physical robot. That is why more and more studies use synthetic data to train imitation learning methods in a virtual environment \cite{finn2017one,yang2014unified,kumar2016learning,zhu2018reinforcement}. 

The purpose of this work is to aid retargeting in the imitation learning pipeline to achieve dexterous tasks using a five-fingered anthropomorphic hand model. In contrast to previous work that used specialised mocap hardware to capture expert demonstrations~\cite{kumar2016learning}, we use a state-of-the-art hand pose estimator (HPE) that aims to extract the skeleton of the hand in every frame captured by a depth camera. These demonstrations are then retargeted to the hand model's space, since that is the environment in which the imitation will take place. Final goal is for the agent to learn to perform the same tasks in its own environment using its own architecture.

The major problem in this pipeline that this work aims to address is the difficulties in successfully retargeting human motion, especially in real-time. These arise from the discrepancies between the human hand and the hand model, the different constraints of the environments and the noise coming from the camera and the hand pose estimator, whose output is less accurate when unseen views and articulations are captured. This leads to an output that is often not kinematically plausible. Hand pose estimators are also vulnerable to occlusions~\cite{yuan2018depth}, be it self-occlusions~\cite{Ye_2018_ECCV,jang20153d} or occlusions from object manipulation \cite{rogez20143d,garcia2018first}. All these discrepancies become especially important when the task involves delicate interaction with other objects, like grasping. In these cases, even the smallest error in the position of the end effectors can result in failure of the task. 

\begin{figure}[ht]
\centering
\includegraphics[trim={1cm 7cm 3cm 4cm},width=1.02\textwidth]{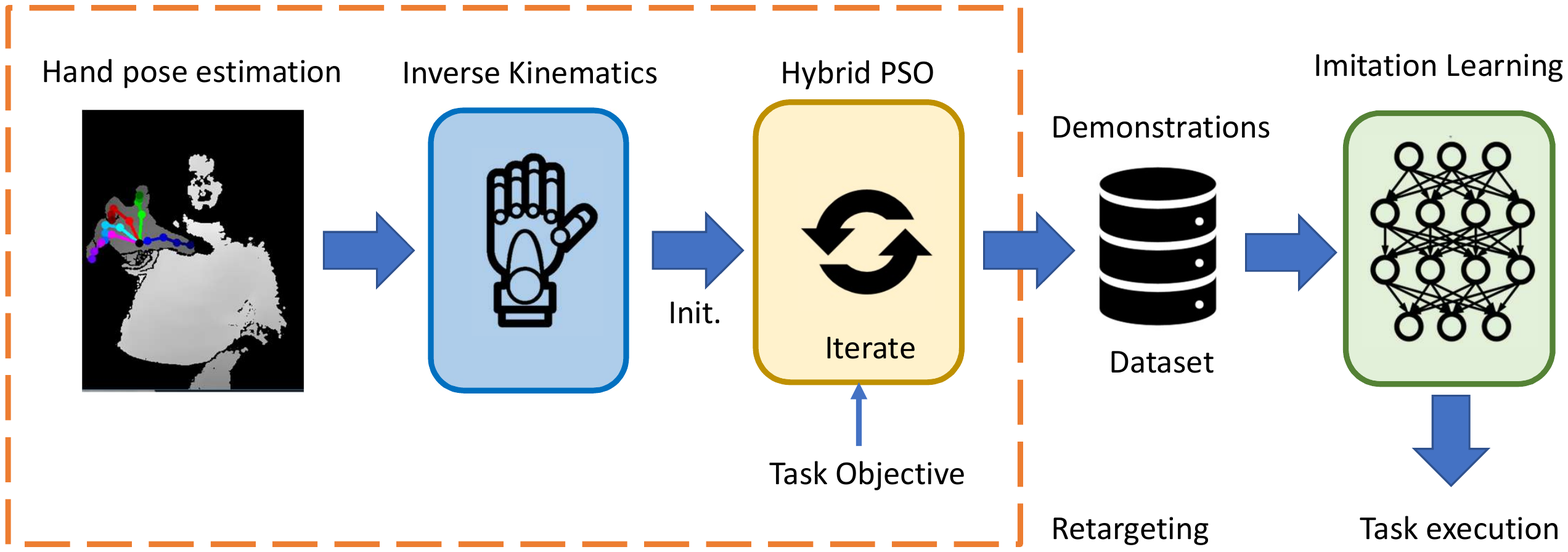}
\caption{Flow chart of our proposed framework.}
\label{fig:system}
\end{figure}

This work addresses this difficulty by introducing task objective optimisation through the use of the particle swarm optimisation (PSO) algorithm \cite{kennedy1995j}. The aim of this is to improve interaction with objects in the agent's environment, which allows easier acquisition of expert demonstrations for imitation learning. In order to achieve more robust results, we developed a Hybrid PSO method that uses inverse kinematics (IK) retargeting for pose initialisation and then refined that pose by driving it to touch the objects with as many fingers as possible. Recorded trajectories are then used to teach an agent how to grasp in a simulated environment by using a generative adversarial imitation learning (GAIL) \cite{ho2016generative}. An overview of our framework is depicted in  Figure~\ref{fig:system}.
\section{Related Work}
\textbf{Hand motion retargeting} to anthropomorphic robot grippers is challenging to control due to their high dimensionality and kinematics discrepancy. Similar to our inverse kinematics baseline, \cite{li1989grasping} proposed to use relative transformations of positions, velocities and forces for grasping. As a result of advances in processing capacity, an interest for simulation environments for robotic learning and for virtual/augmented reality (VR/AR) has emerged. 
In robotic learning, simulation is usually combined with imitation learning~\cite{finn2017one,zhang2017deep}, reinforcement learning~\cite{devin2017learning}, or a combination of the two~\cite{nair2017overcoming,zhu2018reinforcement}. In all those cases however, the used grippers are of low dimensionality (i.e. two or three moving joints) with the exception of a few works that we review in the next section. Regarding manipulation in a VR/AR, \cite{borst2005realistic} and \cite{buchmann2004fingartips} used a bespoke hand glove and haptic sensors to capture hand information and mapped this information to either a realistic spring hand model or a simplistic two fingertips model. Physics-based approaches modelling hand meshes and forces for object interaction include \cite{zhao2013robust,kim2015physics,holl2018efficient}. \cite{holl2018efficient} chose to use a spring hand model to infer contact forces, whereas \cite{kim2015physics} optimised hand deformation using a particle based approach. Similar to our work but requiring a large prerecorded motion dataset, \cite{zhao2013robust} aimed to minimise the distance between contact points and the object in a data-driven fashion through the use of PSO.

In the hand pose estimation field, the interaction with (real) objects has also been studied~\cite{hamer2010object,romero2010hands,Oikonomidis2011,rogez20143d,mueller2017real,garcia2018first,Tzionas:IJCV:2016}. Similar to our work but with a different objective, \cite{Tzionas:IJCV:2016} used physics simulation and salient points to accurately estimate the hand pose in an object manipulation scenario. In our work, we circumvent the object occlusion by working in a mixed reality scenario similar to \cite{mueller2017real}.

To conclude, we include related work on human body motion retargeting. \cite{VNect_SIGGRAPH2017} and \cite{villegas2018neural} used deep learning techniques in a cyclic manner to minimise the error between angles and different models respectively. More related to our work, given their aim to accomplish a task, \cite{2018-TOG-DeepMimic} combined IL and reinforcement learning to both perform tasks and achieve retargeting of motion to different models. Similarly to us, their reward function aims to minimise the distance in position and rotation between the expert demonstrations and the generated sequences.

\textbf{Imitation learning (IL)}, or learning from demonstrations \cite{schaal1997learning} is the process through which, given a set of demonstrations, an agent learns to map observations to actions \cite{hussein2017imitation}. The field has not been a exception of the success of deep learning approaches \cite{hussein2017imitation,finn2017one,duan2017one,ho2016generative}. A recent successful approach is generative adversarial imitation learning \cite{ho2016generative} (GAIL). In their work, \cite{ho2016generative} designed a generative adversarial system inspired by the success of generative adversarial networks (GAN) \cite{goodfellow2014generative}. \cite{ho2016generative} adapted the GANs concept of adversarial behaviour to IL as a policy search problem, where the generator aims to produce the ideal policy that will produce results similar to the demonstrations. Related to our framework, \cite{kumar2016learning} use a mocap system consisting of a glove to record demonstrations in a virtual space that are late fed to an IL system. The key difference between their work and ours is that our input data is noisy due to the hand pose estimation stage and thus we propose a system that can help to alleviate the challenge of recording successful demonstrations.

\section{Framework Overview}
In our framework (Figure~\ref{fig:system}), we use the hand skeleton, provided by the HPE, as input and retarget it to a hand model in a virtual scene. Retargeting is performed, first by approximating the pose with inverse kinematics (IK), and then optimising that pose for a number of iterations. The aim of the optimisation is to grasp an object, thus making lifting easier. Finally, this retargeting method is used to record multiple demonstrations of object lifting, which are then used to train an imitation learning network using the GAIL methodology. Final outcome is an agent that can imitate grasping and lifting objects with a dexterous hand model.

\begin{figure}[ht]
\centering
\includegraphics[width=\linewidth]{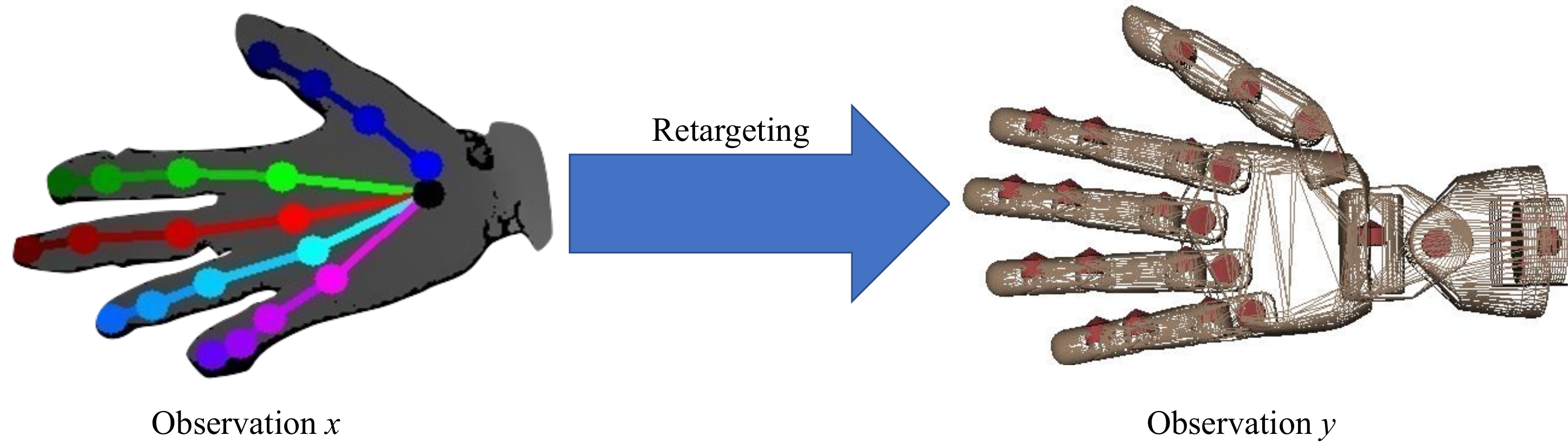}
\caption{Left: the input skeleton points from the hand pose estimator. Right: the hand model that performs imitation and its 23 internal actuators that control joint rotations.}
\label{fig:model}
\end{figure}

The hand pose estimator (HPE) used provides the 3D locations of 21 skeleton points of the hand. In order to perform dexterous actions, the hand model needs to be five-fingered and similar to the human hand. The hand model selected is a modified version of the MPL model presented in \cite{kumar2015mujoco}, which in turn is a virtual model of the Modular Prosthetic Limb developed by Johns Hopkins University \cite{mcgee2014demonstration}. The hand pose can be controlled through 23 actuators, each of which can rotate a joint by a given axis. Therefore, each actuator has 1 degree of freedom. Aside from the hand pose actuators, the hand model also has 6 degrees of freedom for its global position and rotation. Therefore, the model has a total of 29 degrees of freedom (Figure~\ref{fig:model}).

Since the objective of this work is to retarget the hand pose to perform a task, the fitness function of the PSO is selected to minimise the difference between the input observation \textit{x}, which is the skeleton produced by the HPE, and the observation \textit{y} produced by the virtual environment after a particle had been applied to the simulation. Observations $x$ and $y$ can be seen in Figure~\ref{fig:model}.

\clearpage

\section{Retargeting Through Stochastic Optimisation}
In our framework, the mapping process of the hand pose in the real world to the corresponding action space in the virtual environment is achieved through the use of the particle swarm optimisation method \cite{kennedy1995j}, which is a type of stochastic optimisation. In PSO, a number of particles try to find an optimal solution by moving in the search space. Their position is evaluated by a fitness function and their new position depends on their personal and global best. As presented in \cite{shi1998modified}, for each particle \(i\), its position \(p\) and velocity \(v\) for iteration \(t\) are given by:
\begin{align}
v_{i,t} &= v_{i,t-1} + c_1 * (p_{i, t-1}^{best} - p_{i, t-1}) + c_2 * (g_{t-1}^{best} - p_{i, t-1}),\\
p_{i,t} &= p_{i,t-1} * v_{i,t},
\end{align}
where \(p_{i, t-1}^{best}\) is the current personal best position of the particle, \(g_{t-1}^{best}\) is the current global best position for all particles and \(c_1\) and \(c_2\) are learning rates.
\subsection{PSO Fitness Function}
\begin{figure}[ht]
\centering
\includegraphics[width=0.6\linewidth]{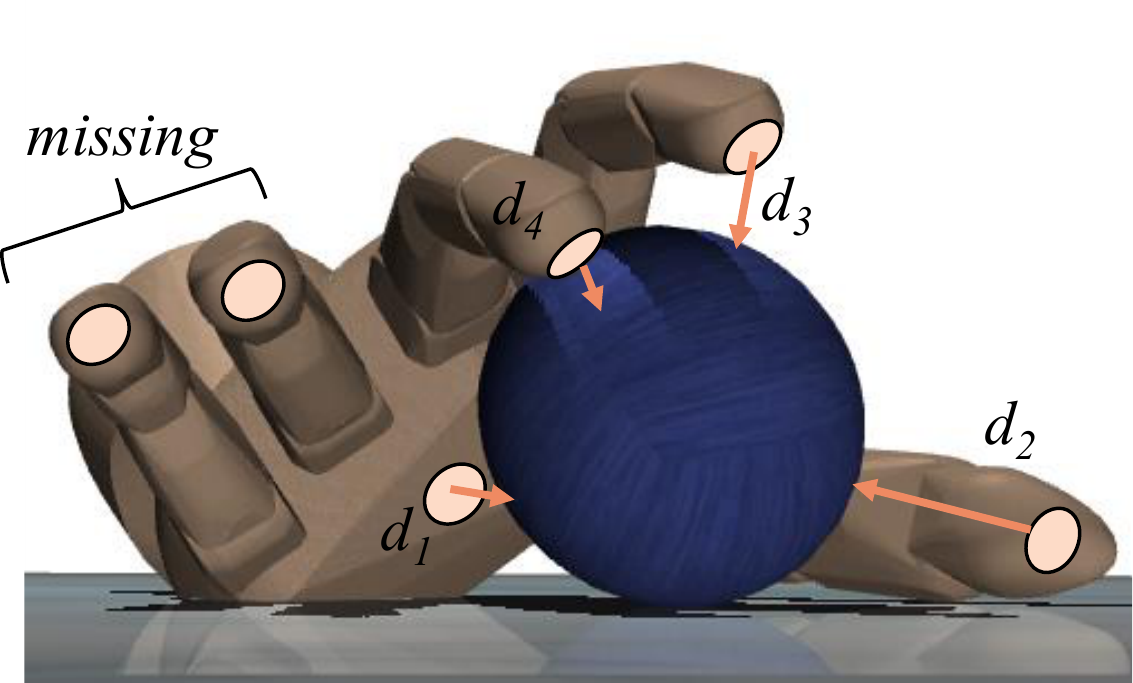}
\caption{Representation of the 6 points of the task energy function. When a point is close to the object, the minimum distance between the point and the object surface is used. When a point is too far from the object to achieve contact, the point is flagged as missing}
\label{fig:contact}
\end{figure}

A key component in PSO is accurately defining its fitness function. In our work, we define the particles and the fitness function in different domains and involve a simulation integration between them. Whereas the particles are the actions that control the hand model, the fitness function evaluates the resulting hand pose of those actions. Therefore, the fitness function does not interact with the positions of the particles themselves, but rather with the outcome of those positions when they are applied to the environment. That way the fitness function aims to infer not only the effect the actions have on the hand pose, but also characteristics of the environment itself (i.e. external forces like gravity, impediment from other objects like walls, collisions).
\\ \\
Aside from matching the reference pose of the HPE skeleton, the PSO can also be used to aid in task execution. Hence, the final form of the fitness function is the following:
\begin{equation}
E(x,y,object) =  \omega_{pose}E_{pose}(x,y) + \omega_{task}E_{task}(y,object),
\end{equation}
where \(E_{pose}\) and \(E_{task}\) represent the hand pose approximation and the task objective respectively, and \(\omega_{pose}\) and \(\omega_{task}\) are their respective weights.
\\ \\
\textbf{Hand Pose Energy Function $E_{pose}$ }
\\ \\
The \(E_{pose}\) energy function minimises the pose difference between input $x$ and output $y$. Similarly to us, \cite{tompson2014real} and \cite{Makris2015} used PSO to minimise the distance between the joints. In our work, though, we are also interested in the relative joint angles. \cite{2018-TOG-DeepMimic} also minimised position and rotation of joints in their reward function, but they were only interested in the end effectors, whereas we want to optimise all the controls. Therefore, the hand pose energy function is:
\begin{equation}
E_{pose}=\omega_pE_p+\omega_aE_a.
\end{equation}
\begin{figure}[ht]
\centering
\includegraphics[width=1\linewidth]{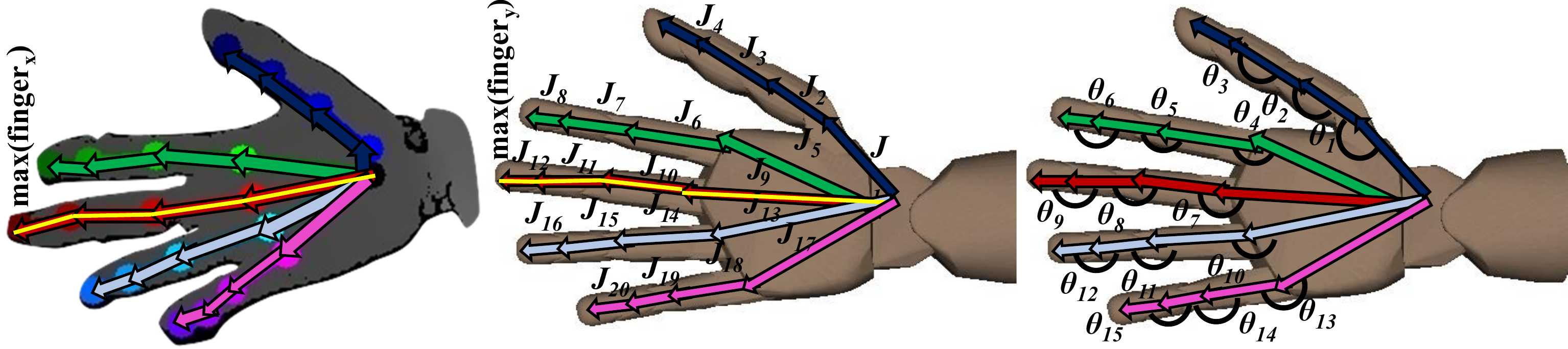}
\caption{Left: Joint vectors J in observation \textit{x}. Middle: Joint vectors J in observation \textit{y}. Right: Angles \(\theta\) in observation \textit{y}}
\label{fig:fitness}
\end{figure}
\paragraph{Position Energy Function $E_p$:} it aims to minimise the euclidean distance between the corresponding points of observations \textit{x} and \textit{y}, and it is set as the normalised weighted mean squared error between them:
\begin{equation}
E_p = \dfrac{1}{\sum_{i=1}^{N_{joints}}\omega_{joint}^{(i)}}\sum_{i=1}^{N_{joints}}\omega_{joint}^{(i)}\|\frac{x\prime(i) - y(i)}{\|\max(finger_{x\prime})\| + \|\max(finger_y)\|}\|^2,
\end{equation}
where $x\prime$ is the representation of $x$ in the $y$ domain. The weights \(\omega_{joint}\) represent the individual weights of each joint allowing prioritisation of some joints, such as the end effectors, above others. $N_{joints}$ is the number of joints, being 21 in our model.
\\ \\
Since the two observations exist in different domains, observation \textit{x} needs to be scaled to the environment and the hand model of observation \textit{y}. The scaling factor used is the global ratio between the joint lengths of the two hand models. Therefore, if the joint vectors of domains $x$ and $y$ are $J_1(x)\ldots J_{N_{joints}}(x)$ and $J_1(y)\ldots J_{N_{joints}}(y)$ respectively (Figure~\ref{fig:fitness}), then the normalised observation \textit{x} is:
\begin{equation}
x\prime = sx \textnormal{ where }s = \dfrac{1}{N_{joints}-1}\sum_{i=1}^{N_{joints}-1}\dfrac{\|J_i(y)\|}{\|J_i(x)\|}.
\end{equation}
\paragraph{Angle Energy Function $E_a$:} While the position energy function minimises the distance between the points, thus preserving the position of the end effectors as well as the general structure and orientation of the hand pose, it is not guaranteed it can accurately preserve the local angles of all 21 points. Therefore, a second energy function \(E_a\) is introduced, that aims to minimise the difference between all the relative angles of the joints, $N_{angles}$, 15 in our model. For \(\theta_i(x)\) and \(\theta_i(y)\) as the 3D joint angles of observation \textit{x} and \textit{y} respectively (Figure~\ref{fig:fitness}), the angle energy function is the normalised mean squared error between the two observations:
\begin{equation}
E_a = \dfrac{1}{N_{angles}}\sum_{i=1}^{N_{angles}}\|\dfrac{\theta_i(x) - \theta_i(y)}{\pi}\|^2.
\end{equation}
\\ \\
\noindent\textbf{Task Energy Function $E_{task}$}
\\ \\
The second term of the fitness function \(E_{task}\) aims to assist in achieving the task's objective. Since the current task is grasping, the intention of this energy function is to minimise the distance between specific points of the hand model and the object in question. The points selected were the middle of the palm and the end effectors, which are the five fingertips (Figure~\ref{fig:contact}).
With $\omega_{palm}$ and $\omega_{ee}$ as the individual weights of the palm and the end effectors respectively, the task energy function is the following:
\begin{equation}
E_{task} = \dfrac{1}{5\omega_{ee} + \omega_{palm}}(\sum_{i=2}^6\omega_{ee}\|\dfrac{d_i}{\omega_{cost}d_{max}}\|^2 + \omega_{palm}\|\dfrac{d_1}{\omega_{cost}d_{max}}\|^2),
\end{equation}
where $d_i$ is the minimum distance between point $i$ and the various meshes of the object. To limit the distance search space, when a point's distance to the object is over a certain threshold $d_{max}$, then it is presumed it cannot interact with the object and is flagged as missing. Therefore, each point's distance to the object is:

\begin{equation}
d_i =\begin{cases}
 min(d_{ij}) & \text{if} \ d_{ij} < d_{max} \\
 \omega_{cost} d_{max} & \text{otherwise}
\end{cases} \quad \text{,}
\end{equation}
where \textit{i} are the six points, \textit{j} are all the meshes of the object and \(d_{ij}\) is the minimum planar distance between the two (Figure~\ref{fig:contact}). The points flagged as missing are punished by assigning them the maximum detectable distance multiplied by \(\omega_{cost}\). The purpose of the weight is to act as an incentive to have as many points close to the object as possible.

\subsection{Hybrid PSO - Task Optimisation}

Due to the stochastic nature of PSO, it is not certain that the algorithm can manage to converge to an acceptable degree, especially when it needs to perform in real-time. For that reason, a Hybrid PSO method was developed, which uses an inverse kinematics approximation as a priori knowledge and applies localised PSO around that pose for task optimisation. This is achieved mainly thanks to the task objective energy function \(E_{task}\) that encourages contact.

The final Hybrid PSO method with all the refinements starts by initialising all the particles in a position similar to the one provided by the inverse kinematics approximation. Then, through a limited number of iterations, it updates the position of the particles, but only up to a certain amount. Finally, the algorithm terminates when a particle converges to an adequate degree. Even if the particles do not have enough time to reach that point due to execution time constraints, the result is still guaranteed to be close to the initial inverse kinematics pose and, therefore, stochasticity from the PSO is constrained.
{\footnotesize
\begin{algorithm}[ht]

\SetKwInput{KwData}{Input}
\SetKwInput{KwResult}{Output}
\SetKwComment{Comment}{}{}
\DontPrintSemicolon
\SetAlgoLined
\KwData{x observation from HPE, swarmsize, iterations}
\KwResult{\(particle_{global}^{best}\)}
 actions = IK(x)\;
 \Comment*[l]{// initialise position of particles around the IK position}
 \For{particle in swarmsize} {
 $particle = actions + rand()$\;
 $particle_{personal}^{best} = particle$\;
 }
 $particle_{global}^{best} = particle$ which has the minimum $E(x, y,contact)$\;
 \BlankLine
 \Comment*[l]{// optimise particles by minimising their fitness function}
 \For{t \textup{\textbf{in}} iterations}{
  \For{particle \textup{\textbf{in}} swarmsize}{
    Update $particle$ position and velocity according to PSO algorithm\;
	$y_{particle}, contact = simulation(particle)$\;
    $E_{t, particle}(x, y, object) = \omega_{pose}E_{pose}(x,y) + \omega_{task}E_{task}(y, object)$\;
    \If{$E_{t,particle}(x,y,object) < E_{personal}^{best}$}{
    	$E_{personal}^{best} = E_{t, particle}(x, y, contact)$\;
    }
   }
   \If{$min(E_{t}) < E_{global}^{best}$}{
   		$E_{global}^{best} = min(E_{t})$\;
    }
 }
 \caption{Proposed Hybrid PSO Algorithm}
 \label{algo:hpso}
\end{algorithm}
}

An additional main advantage of our method is that Hybrid PSO can overcome the HPE's limited response time. Whereas retargeting with the pose energy function $E_{pose}$ solely depends on observation $x$ that comes from the HPE, the task energy function $E_{task}$ only depends on the current state of the environment (observation $y$ of the hand model and the object). That means that it can be applied at a higher frame rate than the HPE, providing continuous micro-corrections while the new hand pose is estimated. It can, therefore, lift the interaction limitations that stem from HPE's bottleneck. Algorithm~\ref{algo:hpso} shows in detail the integration of our Hybrid method in the PSO pipeline, presented in \cite{shi1998modified}.

\section{Imitation Learning}
To complete the framework's pipeline, we recorded a dataset of 161 grasping successful demonstrations using our Hybrid PSO retargeting method and fed them to a GAIL framework. The network aims to learn a policy between states and actions. The state space used in this framework comprises the relative distance and velocity between the hand model and the object, the local rotations and velocities of the internal hand joints, as well as the minimum contact distance between the hand end-effectors (five fingertips) and the object.
\begin{equation}
s_t = [pos_{hand} - pos_{object}, vel_{hand}-vel_{object}, angle_{joints}, vel_{joints}, d_{contact}].
\end{equation}
The action space consists of the 29 actuators of the hand model, which denote its 29 degrees of freedom.

\section{Experiments}
This section describes the experiments conducted to evaluate the effectiveness of the retargeting optimisation. First, the optimisation was tested under retargeting generalisation, where retargeting was performed solely using the PSO approach, without any prior information. Second, the hybrid method was tested in its ability to grasp compared to the inverse kinematics baseline. Third, the GAIL network trained with Hybrid PSO demonstrations was tested in its ability to imitate grasping.
\\ \\
The simulation environment used in this work was Mujoco Pro \cite{kumar2015mujoco} and the hand pose estimator was the one presented in \cite{ye2016spatial} and trained with BigHand2.2M dataset \cite{yuan2017bighand2}. All the experiments were performed using an Intel Core i5-7600K 3.8GHz CPU, an  NVIDIA GeForce GTX 1080 Ti GPU and 32 GB of RAM.
\\ \\
All the results presented used $\omega_{joint}=10$ for the thumb fingertip, $\omega_{joint}=3$ for the remaining fingertips, $\omega_{joint}=1$ for the rest of the joints, $\omega_{palm}=3$, $\omega_{ee}=1$, $\omega_{cost} =2$, $d_{max}=0.04$ and PSO minimum fitness step $= 10^{-4}$.
\subsection{Retargeting Generalisation}
The pose energy function $E_{pose}$ was evaluated through the application of PSO, excluding the task energy function and comparing the results with the input hand pose observation $x$.
\begin{figure}[ht]
\centering
\includegraphics[width=1\linewidth]{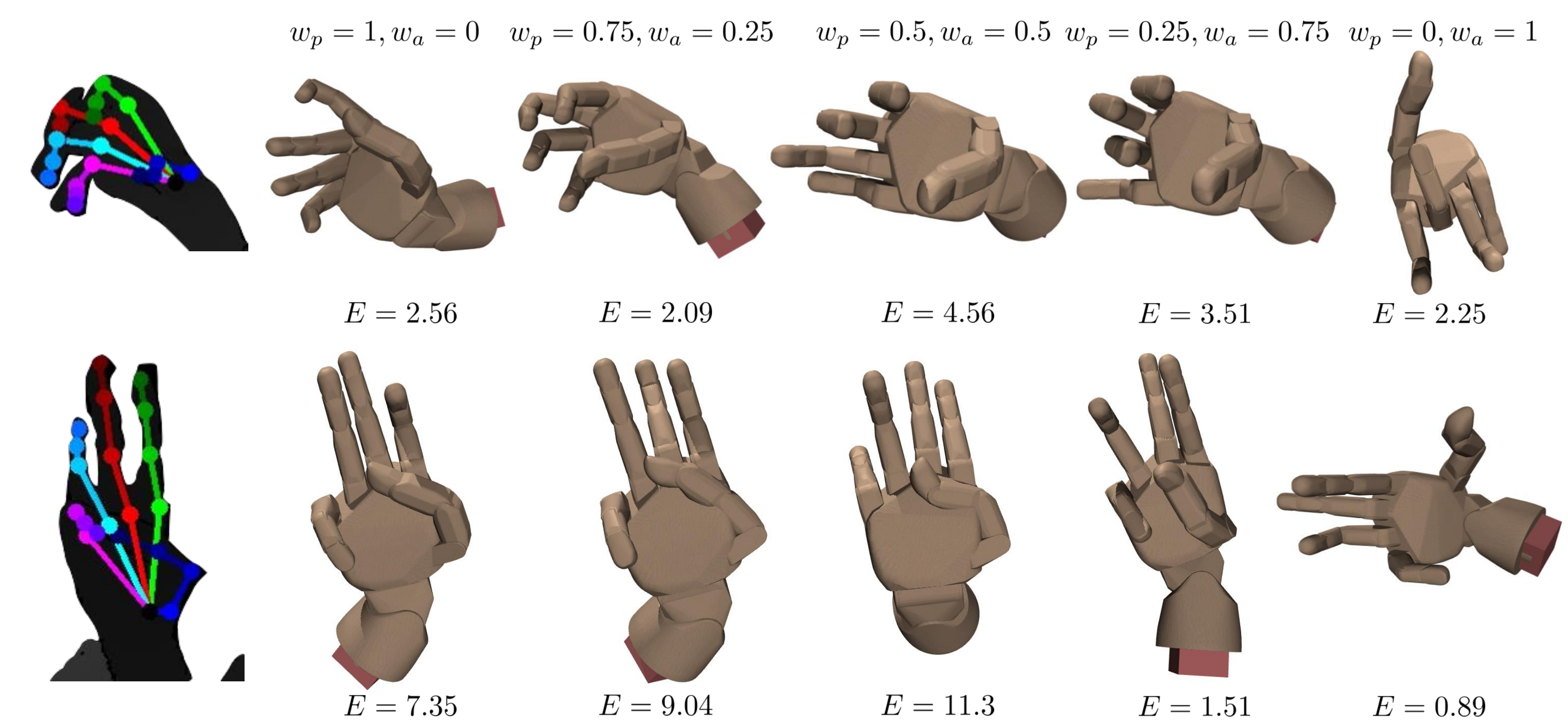}
\caption{Generalised PSO with different weights between position and angle terms (for illustration purposes energy values are scaled by a factor of $10^{3}$).}
\label{fig:pso_angle}
\end{figure}
Figure~\ref{fig:pso_angle} shows the retargeting results using various combinations of the position and angle terms. The position energy function is the most important one since that is the one responsible for the position of the end effectors as well as the global rotation of the hand. The angle energy function on the other hand aims to preserve the relative angles between the joints, so that the final pose is similar to that extracted from the HPE. Using a greater weight for the angle term \(\omega_a\) compared to the position term weight \(\omega_p\) can lead to erroneous rotation directions, as evident in Figure~\ref{fig:pso_angle}. For accurate retargeting results, the position term needs to be the most significant, with the angle energy function acting as auxiliary. That is because the angles do not hold the 3D direction (i.e. the rightmost images of Figure~\ref{fig:pso_angle}, where global hand rotation is not represented and the finger angles are correct in magnitude but can be wrong in direction). That error can be alleviated with the position energy function $E_p$, which minimises the position error of the points. On the other hand, using only $E_p$ does not guarantee the overall pose will be preserved as it is susceptible to accumulating errors, as seen in the leftmost simulation images of Figure~\ref{fig:pso_angle}. Therefore, finding a balance between the two terms is not trivial. This indicates that the definition of the pose energy function $E_{pose}$ for hand pose retargeting can be improved, but that is not the focus of our research, which mainly targets object manipulation.
\subsection{Hybrid PSO}
Our Hybrid PSO method aims to refine the inverse kinematics hand pose in order to achieve grasping. Therefore, the main coefficient of its fitness function is the objective coefficient \(E_{task}\), the goal of which is to minimise the distance between the end effectors and a particular object. 

In order to evaluate the Hybrid PSO retargeting method, we used our inverse kinematics baseline to record 10 trajectories where the grasping motion mostly failed, and then applied our method on the same actions. The demonstrations were recorded with a steady frame rate of 60 frames per second, and Hybrid PSO was applied to all of them off-line. The fitness function weights for all the experiments were: $\omega_{task} = 0.8$, $ \omega_{pose}=0.2$ and $\omega_{p} = 0.5$, $\omega_a = 0.5$.
\begin{figure}[ht]
\centering
\includegraphics[width=\linewidth]{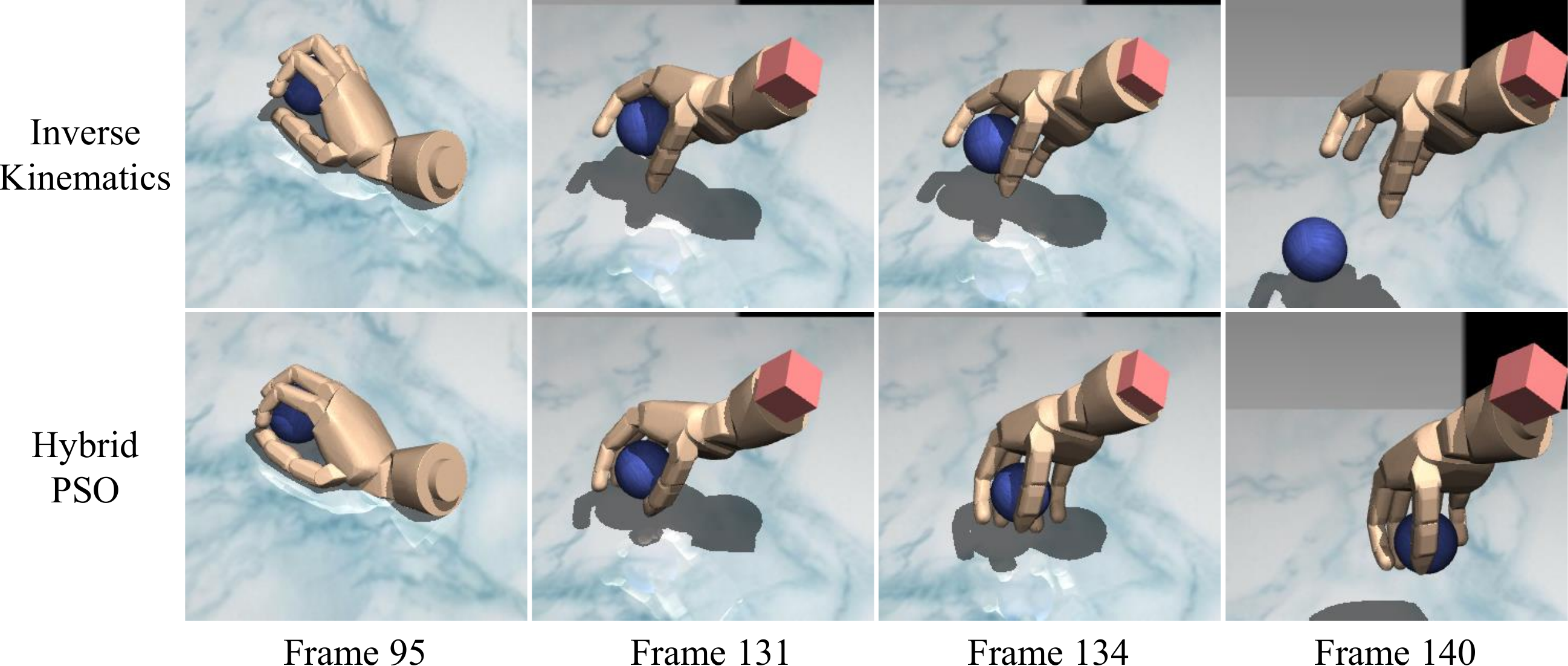}
\caption{Top: Individual frames of trajectory with our inverse kinematics baseline. Bottom: Application of Hybrid PSO on the corresponding frames. The PSO parameters used were: $iterations=100$, $swarm\; size = 100$}
\label{fig:frames}
\end{figure}
Figure~\ref{fig:frames} shows individual frames of a grasping motion with and without Hybrid PSO. This qualitative example indicates that the contact micro-corrections, applied at a high, steady rate, are capable of assisting in the grasping process (full videos at the project webpage).  
\begin{figure}[ht]
\centering
\includegraphics[width=\linewidth]{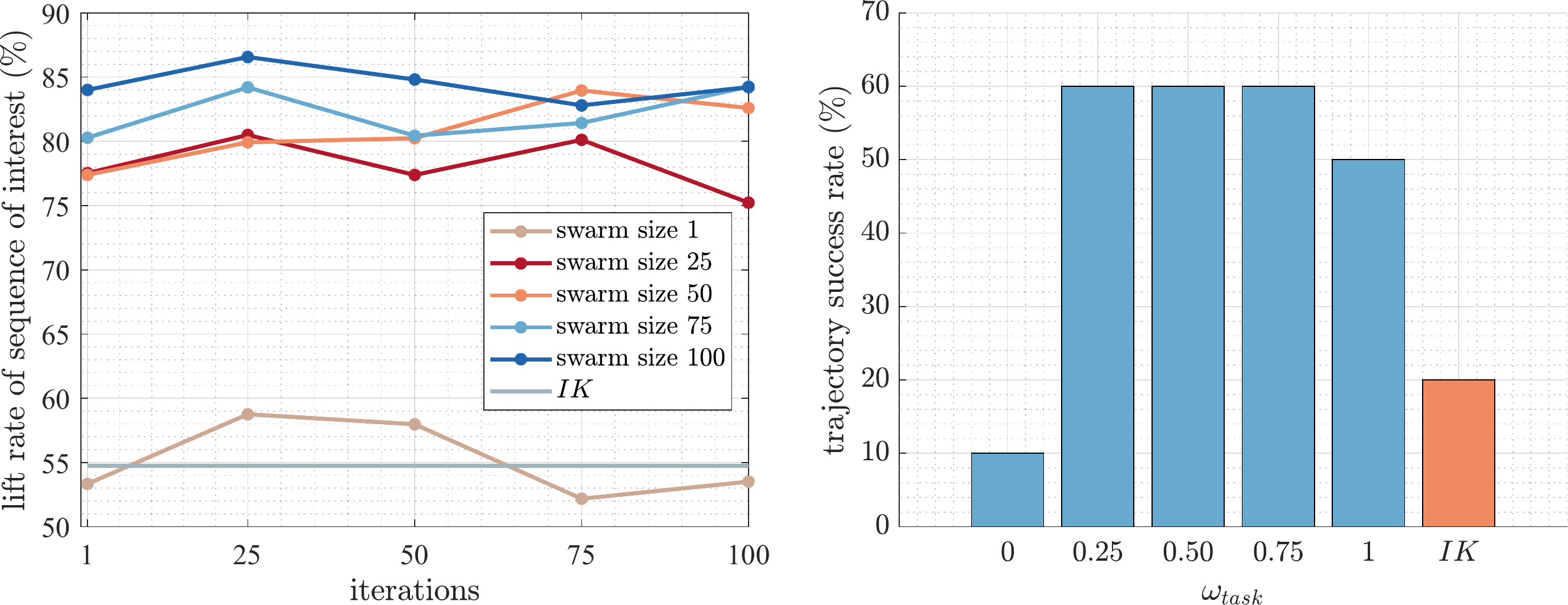}
\caption{Left: Percentage of lifting frames in the ``sequence of interest" part of the trajectories, which is the part where the hand model is capable of lifting the object. Right: Percentage of successful trajectories for different $\omega_{task}$, with $iterations=50$ and $swarm\; size=25$. $\omega_{pose}$ is normalised so that $\omega_{task} + \omega_{pose} = 1$}
\label{fig:success}
\end{figure}

To quantitatively evaluate the method, we measured the percentage of frames the object is being lifted in each sequence. Since the trajectories also include the motion approaching the object, we denote the start of the ``sequence of interest" as soon as at least two of the six points of $E_{task}$ touch the object. From that moment onward, the hand model is in a position that could potentially lift the object. A frame is labelled as successfully lifting only when a) the object is above the table, b) the distance between the object and the palm is less than 0.2 and c) at least one of the six contact points is close to the object. The final lifting ratio is the number of lifting frames divided by the number of frames in the ``sequence of interest". Figure~\ref{fig:success} shows the percentages for different iterations and swarm sizes. It can be seen that a relatively small number of iterations and particles (i.e. 25) can greatly boost the results, compared to our inverse kinematics baseline. It is worth noting that the optimisation quickly saturates, since there is not much improvement between 25 and 100 in swarm size and iterations.

Regarding the importance of the task energy function $E_{task}$, the bar graph on the right of Figure~\ref{fig:success} shows the percentage of successful trajectories for different $\omega_{task}$. A trajectory is classified as successful if the object is being held in the air in the last frame, or if the object is lifted over 17 cm during the sequence. From the graph, it becomes apparent that using only the pose energy function ($\omega_{pose}=1$, $\omega_{task}=0$) produces worse results than the baseline, mainly due to the non-optimal definition of the pose term $E_{pose}$. But the use of greater values of $\omega_{task}$ result in significant improvement of the success ratio, which indicates $E_{task}$ does affect grasping. It should be noted that using $\omega_{task}=1$ produces worse results than other values, which indicates that $E_{pose}$ is also important, since it forces the hand pose to not deviate too far from the $x$ observation and greatly alter the sequence of actions.  
\subsection{Imitation Learning}
The imitation network was evaluated in its ability to successfully grasp from unseen initial conditions. To assist in quicker divergence and more natural motion, the network's generator was pretrained using behavioural cloning (BC). In the GAIL network, the trajectories were initialised using a method similar to \cite{zhu2018reinforcement}, where initial conditions were either extracted from the demonstrations or random, based on a probability \(\epsilon\). The generator and discriminator architectures were identical, with two hidden layers of size 1024 and 512 respectively and \(tanh\) activation functions. Similarly to \cite{ho2016generative}, the generator step function was further optimised using TRPO in order to ensure trajectories would not deviate too much from the demonstrations. The GAIL model was trained for 300 iterations, while its generator was also trained with BC for 100k iterations. We used the implementation of \cite{baselines}.
\begin{figure}[ht]
\centering
\includegraphics[width=0.8\linewidth]{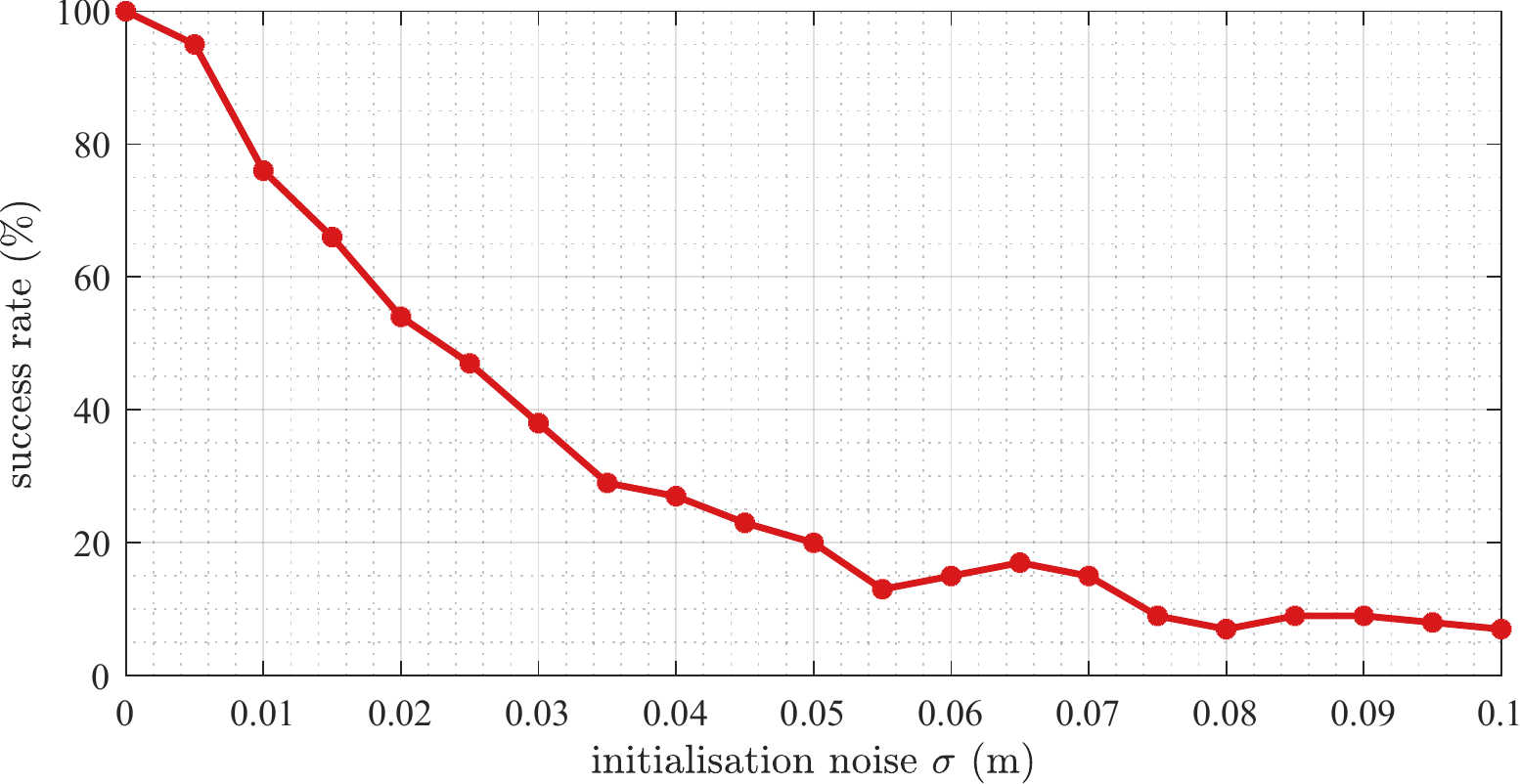}
\caption{Imitation success rate compared to different noise values in the initial state}
\label{fig:sigma}
\end{figure}
The unseen initial conditions were extracted from the demonstrations and then injected with uniform random noise with a standard deviation \(\sigma\). Figure~\ref{fig:sigma} shows the success rate of 100 trajectories compared to different \(\sigma\). It can be seen that, while the policy is accurate when it is initialised at a condition it has previously seen, its success quickly drops when it tries to generalise to similar but unseen initial conditions. That behaviour though is expected, since GAIL offers limited generalisation and multi-modality~\cite{li2017infogail}, which is made even more difficult due to the high dimensionality of this framework's model.
 
\section{Conclusion}
In this work, we presented a Hybrid PSO method, which aims to assist in object grasping, using a hand pose estimator as input. Due to high input noise from hand pose estimators and absence of haptic feedback from virtual objects, the motivation behind this work was to allow easier and more robust grasping using a dexterous, human-like hand model. This method can then be used to record expert demonstrations by directly recording and retargeting human hand motion. Those demonstrations can then be used in a task learning environment, such as imitation or reinforcement learning, and model hand manipulation in a human-like manner. The presented work has the potential to be extended in several ways, such as: i) exploring new tasks and objects via transfer learning; ii) exploiting the hierarchical nature of the hand topology; iii) end-to-end task-oriented retargeting (e.g. reinforcement learning) and iv) enforcing natural hand motion using ground-truth hand pose data of manipulation actions \cite{garcia2018first}.
\\ \\
\textbf{Acknowledgement}. This work is part of Imperial College London-Samsung Research project, supported by Samsung Electronics.

\end{document}